\documentclass[letterpaper, 10 pt, journal, twoside]{ieeetran}
\usepackage{graphicx} 
\usepackage{times} 
\usepackage{amsmath} 
\usepackage{amssymb}  
\usepackage{amsfonts}
\usepackage{array}
\usepackage{algorithm}
\usepackage[noend]{algpseudocode}
\usepackage{multirow,color}
\usepackage{cite}
\usepackage{algpseudocode}
\usepackage{varwidth}
\usepackage{subfig}
\usepackage{booktabs}
\usepackage{gensymb}
\usepackage[hyphens]{url}
\usepackage[export]{adjustbox}
\usepackage[font=footnotesize]{caption}

\usepackage[inline]{enumitem} 
\usepackage{dblfloatfix}
\usepackage{xcolor}
\usepackage{makecell}
\usepackage{textcomp}
\usepackage{gensymb}
\usepackage{amsmath,lipsum}
\usepackage{mathtools}
\usepackage{url}
\usepackage{verbatim}
\usepackage{booktabs}
\begin{document}

\title{KGpose: Keypoint-Graph Driven End-to-End Multi-Object 6D Pose Estimation via Point-Wise Pose Voting}
%
%
%

\author{Andrew~Jeong, and~Seokhwan~Jeong*
\thanks{Andrew Jeong, and Seokhwan Jeong are with Department of Mechanical Engineering, Sogang University, Seoul, South Korea 
.}}


%
%

\markboth{Journal of \LaTeX\ Class Files,~Vol.~XX, No.~X, July~2024}%
{Shell \MakeLowercase{\textit{et al.}}: Bare Demo of IEEEtran.cls for IEEE Journals}
%



\maketitle

\begin{abstract}
This letter presents KGpose, a novel end-to-end framework for 6D pose estimation of multiple objects. Our approach combines keypoint-based method with learnable pose regression through `keypoint-graph', which is a graph representation of the keypoints. KGpose first estimates 3D keypoints for each object using an attentional multi-modal feature fusion of RGB and point cloud features. These keypoints are estimated from each point of point cloud and converted into a graph representation. The network directly regresses 6D pose parameters for each point through a sequence of keypoint-graph embedding and local graph embedding which are designed with graph convolutions, followed by rotation and translation heads. The final pose for each object is selected from the candidates of point-wise predictions. The method achieves competitive results on the benchmark dataset, demonstrating the effectiveness of our model. KGpose enables multi-object pose estimation without requiring an extra localization step, offering a unified and efficient solution for understanding geometric contexts in complex scenes for robotic applications.
\end{abstract}

\begin{IEEEkeywords}
Deep learning, Object pose estimation, End-to-end network
\end{IEEEkeywords}

%
\IEEEpeerreviewmaketitle

\section{Introduction}

\IEEEPARstart{W}{ith} advances in robot technologies, robots are gradually being utilized in industries and in our lives also. In order for robots to accomplish physical tasks effectively and safely along with humans, robots should understand the geometric relationships with their surroundings. One approach to understand it is to estimate 6D pose, which stands for 3D rotation and 3D translation, for each object in the scene and this information is crucial for robotic manipulation or grasping.

\begin{figure}[t!]
    \centering
    \includegraphics[width=0.9\linewidth]{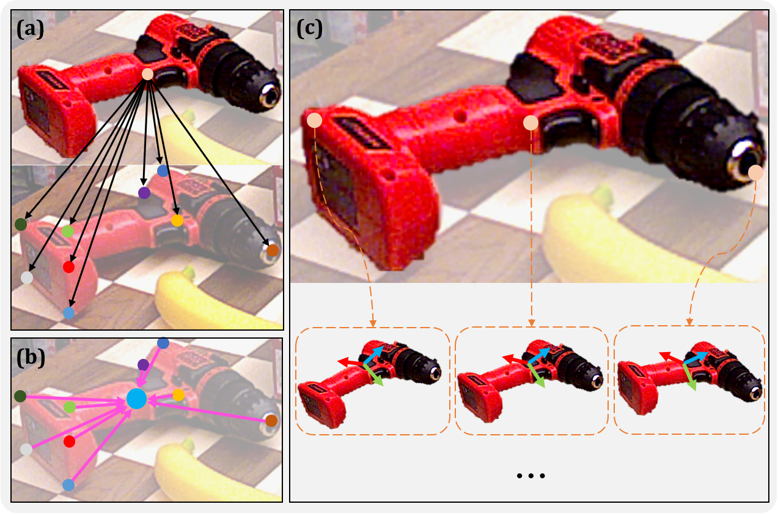}
    \caption{Our approach. (a) Each point in input point cloud votes for 3D keypoints of each object in the scene. (b) Each set of the keypoints are converted into a graph, which is called `keypoint-graph'. (c) 6D pose of an object is estimated (or voted) from each point through several layers of graph convolution.}
    \label{fig:Method digest}
    \vspace{-0.3cm}
\end{figure}

\begin{figure*}[t!]
    \centering
    \includegraphics[width=\linewidth]{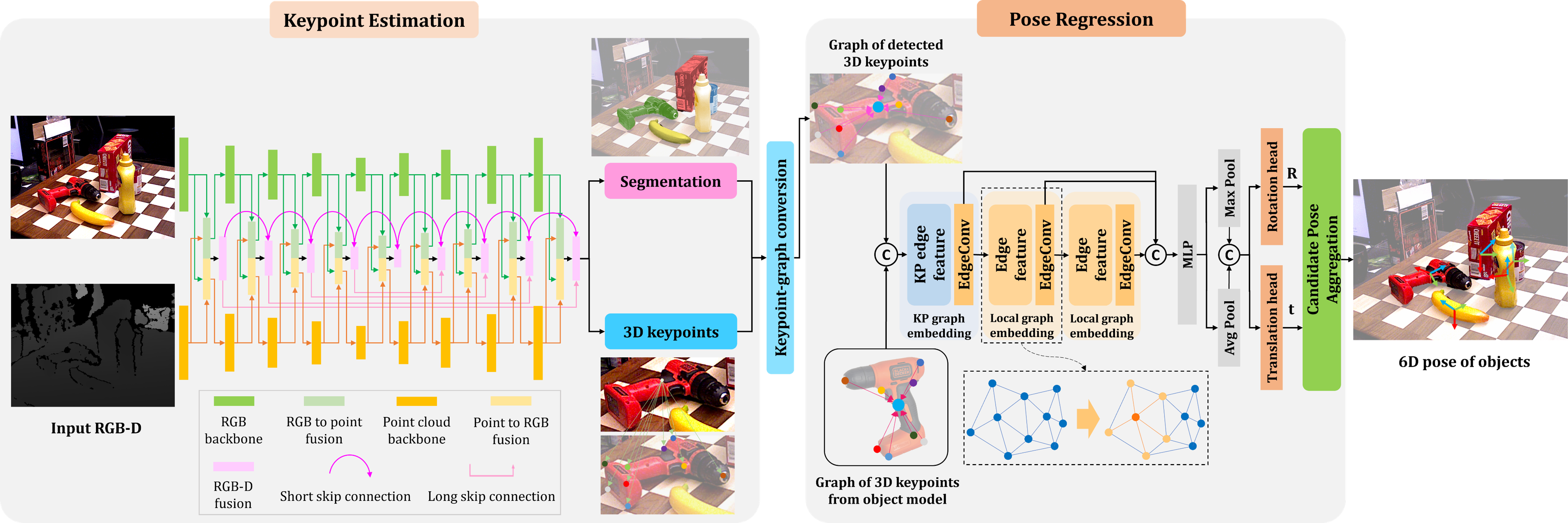}
    \caption{Overview of KGpose. Given RGB-D image, 3D keypoints are estimated through RGB and point cloud branches along with feature fusion process. Then, the estimated keypoints and corresponding keypoints from object model are converted into a graph representation and concatenated. These graphs are passed through keypoint graph embedding and several layers of local graph embedding stages to embed the graph features into each node (or point). Finally the embedded point features are fed into rotation and translation head to regress 6D pose parameters for each point, which are candidates of 6D pose. The final 6D pose for each object is determined by selecting the nearest candidate to the mean of the candidates of the object.}
    \label{fig:Framework}
    \vspace{-0.3cm}
\end{figure*}

\par Common approach for recovering 6D pose of an object is to regress 6D pose from keypoints of each object in the scene. Keypoint-based methods usually consist of two stages; first, estimate keypoints which could be considered as geometric clues for predicting object pose, then predict 6D pose parameters leveraging the estimated keypoints.
Some methods detect 2D locations of the keypoints for each object and estimate 6D pose through RANSAC (RANdom SAmple Consensus) based Perspective-n-Point (PnP) algorithm establishing 2D-3D correspondences between the keypoints in the image and the ones from 3D object model\cite{peng2019pvnet, li2019cdpn, song2020hybridpose}. 
Other methods detect 3D keypoints rather than 2D ones. Estimating 3D keypoints could be preferred to avoid errors occurred from 2D projection, such as a reprojection error, a chance of overlapping of 2D projections from different keypoints in 3D space, or losing geometric constraints for rigid 3D objects\cite{he2020pvn3d, he2021ffb6d, wu2022vote, zhou2023deep}. These methods usually detect 3D keypoints adopting a voting scheme\cite{qi2019deep}, and determine 6D pose through least-squares fitting algorithm\cite{arun1987least, umeyama1991least} given the detected 3D keypoints in camera coordinate system and corresponding 3D keypoints in the object coordinate system.

\par Recent pose estimation researches have focused on directly regressing 6D pose by designing fully differentiable models, which is referred to as an end-to-end pose estimation\cite{xiang2017posecnn, kehl2017ssd, chen2020end, hu2020single, wang2021gdr, di2021so,lin2022e2ek, cao2022dgecn}. There have been some methods that make RANSAC-PnP be differentiable\cite{chen2020end, hu2020single} to achieve end-to-end learning, while other methods try to regress the 6D pose directly from 2D-3D or 3D-3D correspondences\cite{wang2021gdr, di2021so, lin2022e2ek,cao2022dgecn}.
The output of the model could be 3D rigid transformation matrix\cite{lin2022e2ek, cao2022dgecn}, Euler angles with translation offsets\cite{kehl2017ssd}, or quaternions\cite{xiang2017posecnn}.

\par We propose an end-to-end object 6D pose estimation framework that combines keypoint-based method with learnable pose regression module. While most of keypoint-based 6D pose estimation studies perform keypoint estimation, followed by pose regression based on the estimated keypoints, which is whether differentiable or not, we introduce a voting scheme for 6D pose estimation that generates point-wise pose prediction from input point cloud and learn to aggregate the candidates of 6D pose parameters to select the best one. We leveraged graph convolution on `keypoint-graphs', which are constructed by converting estimated 3D keypoints into a graph representation, followed by disentangled 3D rotation and 3D translation head to regress the pose parameters.







\section{Our approach} \label{sec:KGpose}
\par We propose a framework called KGpose that estimates 3D keypoints of each object and leverages the keypoints as intermediate features of given RGB-D image to regress 6D pose directly in an end-to-end manner. Hence, we will describe a process of 3D keypoint estimation, followed by pose regression method through graph convolution on `keypoint-graphs' which are graph representations of the estimated keypoints, with all processes being differentiable to enable end-to-end training for 6D pose estimation.

\subsection{3D Keypoint Estimation}
\subsubsection{Embedding Multi-modal Features} \label{sec:feature extraction}

\par Given an RGB-D image, we designed overall architecture for feature extraction as an encoder-decoder structure and extracted appearance features from RGB branch, ResNet34\cite{he2016deep} and geometric features from point cloud branch, RandLA-Net\cite{hu2020randla}, and fused the features from both modalities to learn correlation between them, following prior works\cite{he2020pvn3d, he2021ffb6d, zhou2023deep}. Our KGpose performs segmentation and keypoint estimation on input  point cloud, without center point estimation as PVN3D\cite{he2020pvn3d} and FFB6D\cite{he2021ffb6d} did.

\par We adopted attention mechanism along with  bidirectional feature fusion\cite{he2021ffb6d} which introduced point-to-pixel features and pixel-to-point features to exploit context about the other branch of different modality to increase representative power of a model. In order for each modality to understand another modality well (e.g., RGB and point cloud), we introduced attention mechanism to the feature fusion and skip connection process. 
There are candidate attention mechanisms to utilize for attentional feature fusion on each layer in RGB branch and point cloud branch: from conventional Squeeze and Excitation (SE) module\cite{hu2018squeeze}, Convolutional Block Attention Module (CBAM)\cite{woo2018cbam} to transformer based methods like Point Transformer\cite{zhao2021point}, Cross ViT\cite{chen2021crossvit}, or Point-BERT\cite{yu2022point}. 
Among them, we adopted CBAM module for both feature fusion and long / short skip connections, in that it has lightweight design but comparable performance to expensive self-attention based transformer architectures in terms of attentive power on 3D point cloud\cite{qiu2021investigating}. Also, it is maneuverable so that designing spatial and channel attention module for point cloud was simple even though the original architecture was designed for 2D images.

\begin{figure}[b!]
    \centering
    \includegraphics[width=\linewidth]{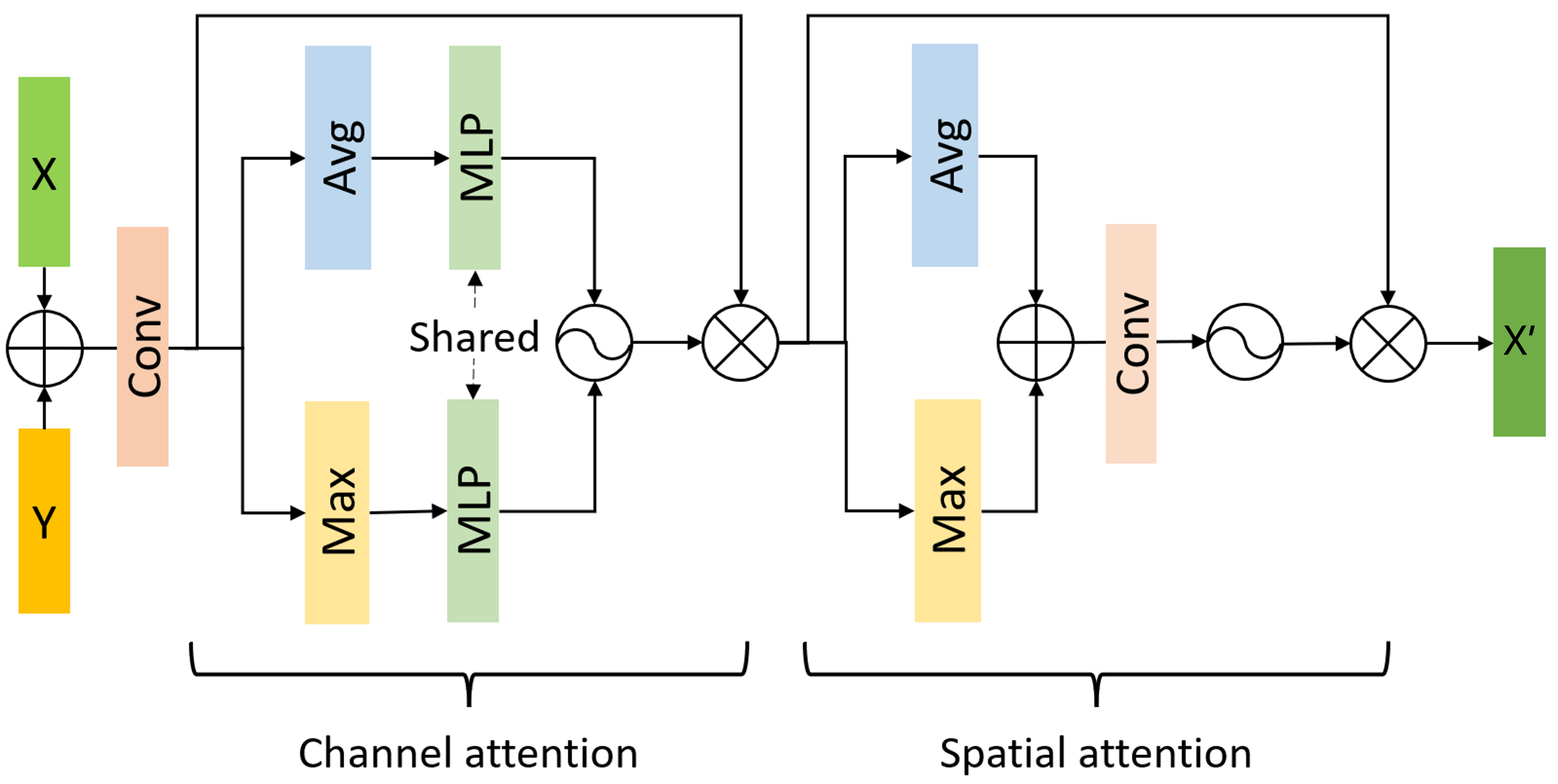}
    \caption{Revised CBAM module for both feature fusion and skip connections.}
    \label{fig:cbam}
    \hspace{0.5cm}
\end{figure}

\begin{figure*}[t!]
    \centering
    \includegraphics[width=\linewidth]{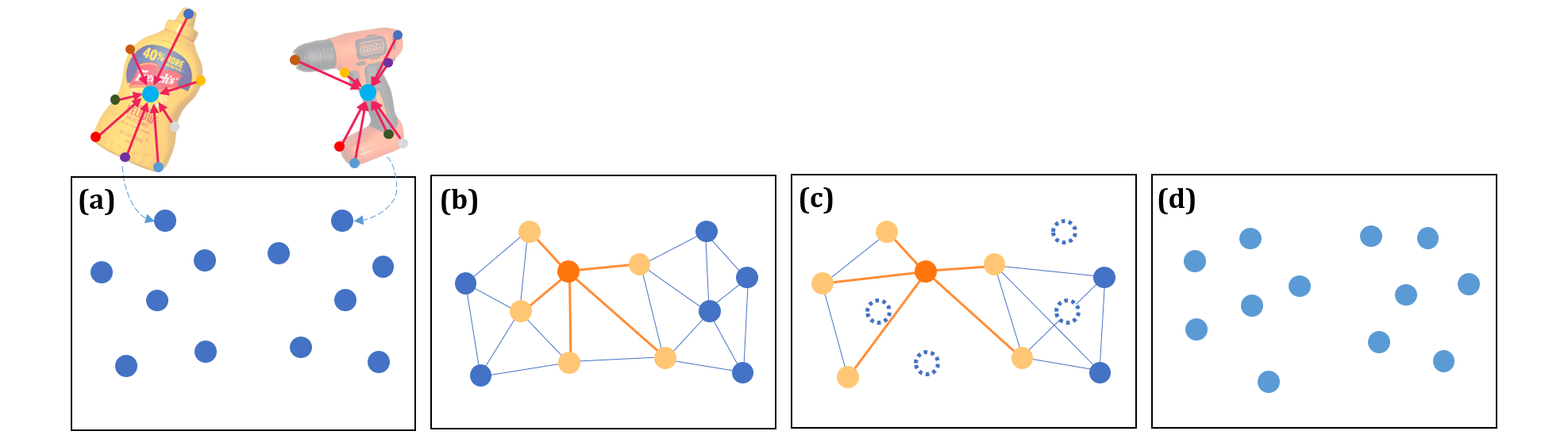}
    \caption{Process of graph embedding. (a) Estimated keypoints are regarded as vertices and edges are defined as vectors from the keypoints to their center. Edge features from each graph are passed through Edge Convolution to embed the information about the keypoints to each point that voted for them. (b) The embedded point features construct local graphs about their k-NN and the edge features from the graphs are also fed into Edge Convolution layer to embed graph features (c) The point features are sampled and construct local graphs about k-NN to understand the features with a larger receptive field (d) The outcomes from (b) and (c) are fused to update the point features}
    \label{fig:graph embedding}
\end{figure*}

\par Attention module was defined as follows:

\begin{align}
\label{eq:attention}
&X'= Mish(X+Att(X, Y)) \\\nonumber
&\text{where}\; \begin{cases}
Att(X, Y) = SA(CA(Conv(X\oplus Y)))\\
CA(X)= \sigma(F(Avg(X)) + F(Max(X))) \odot X \\
SA(X) = \sigma(Conv(Avg(X) \oplus Max(X))) \odot X
\end{cases}
\end{align}

where $Att$ denotes sequential channel/spatial attention module, $CA$ and $SA$ are channel/spatial attention, $Avg$ and $Max$ are average pooling and max pooling respectively, $F$ refers to MLP layers, $BN$ is batch normalization, $Mish$ is Mish activation\cite{misra2019mish}, $\sigma$ is sigmoid and $\oplus$ denotes concatenation. Both input features $X$ and $Y$ are first concatenated, passed through convolution layers, and applied channel attention followed by spatial attention to generate mutual relationship between $X$ and $Y$. Then, feature X is updated by Mish activation following a residual connection.

\subsubsection{Attentional Bidirectional Feature Fusion}
\par We introduced spatial/channel attention to the bidirectional feature fusion. Once the point-to-pixel feature or the pixel-to-point feature are produced following \cite{he2021ffb6d}, these features are fused with RGB feature or point cloud feature with attentional bidirectional feature fusion,  then the outcomes from both modalities are concatenated to produce a fused feature (we would call it `RGB-D feature'), which is different from \cite{he2021ffb6d, zhou2023deep}. The RGB-D feature was concatenated at each layer of the architecture deliberately to facilitate desirable feature extraction on the last RGB-D feature in that segmentation and keypoint estimation are performed on the feature concatenated from the RGB feature and point cloud feature from the last layer in decoder. Referring to \eqref{eq:attention}, $X$, $Y$ are RGB feature and point-to-pixel feature when transferring a context of point cloud feature to RGB branch, and point cloud feature and pixel-to-point feature when transferring RGB context to point cloud branch.

\subsubsection{Attentional Skip-Connection between Fused Features}
\par Each fused feature is derived from the RGB and point cloud features of their respective branches. To effectively capture and understand the fused RGB-D features across different layers, we also incorporated an attention mechanism within the skip connections.

\par As mentioned before, long / short skip connections between fused RGB-D features were also designed with attention modules to help the network focus on the most relevant information when combining the features. Attention modules were applied to RGB-D features for short skip-connections between two consecutive layers and long skip-connections between corresponding layers of the encoder and decoder .

\subsubsection{Estimating Keypoints and Point-wise Semantic Labels}
\par Our KGpose performs 3D keypoint estimation and segmentation on the RGB-D feature learned from both appearance and geometric information. The prediction heads for keypoint estimation and segmentation consist of shared MLPs with different output dimensions.
\par Eight keypoints, representing geometric information for each object, are defined using the SIFT-FPS\cite{he2021ffb6d} and annotated in the datasets. The locations of the keypoints are supervised with two loss functions; $\mathcal{L}_{kp}$ which is L1-norm about 3D keypoint location\cite{he2020pvn3d} and $\mathcal{L}_{geom}$ defined as a sum of cosine similarities about the respective vectors from one keypoint to another to reinforce accurate keypoint estimation guided by their geometric relationships.
\begin{align}
\label{eq:kp loss}
&\mathcal{L}_{kp}=\sum_{c}^C\sum_{n=1}^N\sum_{i=1}^8{|kp_i^n - kp_i^{n*}|}\mathrm{}{I}(\mathcal{C}_n = c) \\
&\mathcal{L}_{geom}=\sum_{c}^C\sum_{n=1}^N\sum_{1\leq i < j \leq 8}\frac{\overrightarrow{kp_{ij}^n}\cdot\overrightarrow{kp_{ij}^{n*}}}{|\overrightarrow{kp_{ij}^n}||\overrightarrow{kp_{ij}^{n*}}|}\mathrm{}{I}(\mathcal{C}_n = c)
\end{align}
where $kp_i^n$ and $kp_i^{n*}$ stands for the prediction and its ground truth about the $i{-th}$ keypoint offset from point $n$, $\overrightarrow{kp_{ij}^n}$  and $\overrightarrow{kp_{ij}^{n*}}$ are those about each vector from $i$-th keypoint to $j$-th keypoint respectively, $\mathrm{I}(\mathcal{C}_n = c)$ indicates whether the semantic label of point $n$ equals to a label $c$.

\par Segmentation is supervised with $\mathcal{L}_{seg}$ defined as Focal Loss\cite{lin2017focal}, which predicts a semantic label for each point of input point cloud.

\subsection{6D Pose Estimation via Keypoint-Graph Representation}

\subsubsection{Keypoint-Graph Embedding}
\par To leverage geometric information about estimated 3D keypoints to learn 6D pose directly, we converted detected 3D keypoints (or offsets from each point) into a graph representation as $\mathcal{G}^K=(\mathcal{P}^K, \mathcal{E}^K)$. $\mathcal{P}^K = \{p^n_i\}$  denotes a set of vertices which are detected 3D keypoints voted from point $p^n$, and $\mathcal{E}^K = p^n_i\rightarrow p^n_c$ are edges about the vertices $p^n_i$ and the center of all keypoints, where $i\in[1, 8]$. Edge features are defined as the concatenated feature of the center of the keypoints and the edge from the keypoints to the center: $e_i=(p_c-p_i)\oplus p_c$ which would have 6 channels.  The keypoints on object model are also converted into graphs in the same way, and the both edge features for the prediction and its corresponding object model are concatenated and passed through Edge Convolution\cite{wang2019dynamic} which is a channel-wise operation on the edge features, followed by max operator to embed the features as point features (see Fig. \ref{fig:graph embedding}(a)). So, the keypoint-graph embedding to produce point feature can be expressed as:
\begin{align}
\label{eq:dkgc}
&{p^{n}}' = \max_{i:(i, c)\in\overline{\mathcal{E}}^K}(Mish(h_{\Theta}(\overline{p}^n_i, \overline{p}^n_c)))\\
& \text{where} \ 
\begin{cases}
h_{\Theta}(\overline{p}^n_i, \overline{p}^n_c) = \theta_i\cdot(\overline{p}^n_i-\overline{p}^n_c)+\phi_i\cdot \overline{p}^n_i\\\nonumber
p^n_c={1\over8}\sum^8_{i=1} p^n_i\\
\overline{p} = p_{pred} \oplus p_{model}.
\end{cases}
\end{align}

\subsubsection{Local Graph Embedding for Learned Point Features}
\par Given the embedded point features, we construct local graph $\mathcal{G}^L=(\mathcal{P}^L, \mathcal{E}^L)$ about k-nearest neighbors (k-NN) for each point in its feature space and perform Edge Convolution to update the point features at each layer, where the operator to generate edge features is defined as asymmetric edge function inspired by DGCNN\cite{wang2019dynamic} (see Fig. \ref{fig:graph embedding}(b)). 
\begin{align}
&x'_i=\max_{j:(i,j)\in\mathcal{E}^l}(Mish(f_{\Theta}(x_i, x_j))) \\\nonumber
&\text{where} \ f_{\Theta}(x_i, x_j)=\theta_i\cdot(x_i-x_j)+\phi_i\cdot x_i\\\nonumber
\end{align}
\par Upon this procedure, we suggest to random sample the point features (or vertices) and construct another local graph $\mathcal{G}^l=(\mathcal{P}^l, \mathcal{E}^l)$ by gathering k-NN over sparse point features (see Fig. \ref{fig:graph embedding}(c)). The point features are upsampled and the outcomes from both processes are fused to update the point features capturing a larger receptive field over the point features (see Fig. \ref{fig:graph embedding}(d)).
\begin{align}
&({x'_i}^l)_{up} = Upsample({x'_i}^l) \\\nonumber
&x'_i = Conv({x'_i}^L \oplus ({x'_i}^l)_{up}) \\\nonumber
\end{align}
\par The graphs are updated several times and the outputs from every layers are aggregated to regress 6D pose parameters from the set of keypoints that each point of input point cloud voted for. The aggregated features are passed through MLP followed by average pooling and max pooling, then the outcomes are concatenated and fed to rotation head and translation head, each of which consists of several MLPs.

\subsubsection{Regressing 6D Pose from Embedded Features}
\par It is known that all representations for 3D rotation are discontinuous in Euclidean space for four or fewer dimensions, such as 3D Euler angles or 4D quaternions, which causes difficulties for a neural network in learning 3D rotation\cite{zhou2019continuity}.  Instead, a 6D representation $R_{6D}$ for $R\in SO(3)$ was proposed as continuous representation in Euclidean space and shown to have better results on estimating the rotation\cite{zhou2019continuity, wang2021gdr, lin2022e2ek}. The $R_{6D}$ is simply defined as two columns of 3D rotation matrix:
\begin{equation}
    R_{6D}=[r_1|r_2].
\end{equation}
Our network would predict these two columns (or 6 parameters) through rotation head and retrieve rotation matrix $R=[R_1|R_2|R_3]$ according to:
\begin{align}
\begin{cases}
    R_1 = \mathcal{N}(r_1)\\
    R_3 = \mathcal{N}(R_1 \times r_2)\\
    R_2 = R_3 \times R_1
\end{cases}
\end{align}
where $\mathcal{N}$ denotes normalizing operation on a vector. Through this postprocess, the predicted $R_{6D}$ can be converted to 3D rotation matrix without losing its orthonormality.

\par We have candidates for 6D pose as much as the number of input point cloud, as each point predicts (or votes for) a set of 6D pose parameters based on the estimated keypoints given its semantic label and it is pretty similar to a voting scheme for keypoints.

\subsubsection{Selecting 6D Pose of Each Object from Candidates}
\par After performing pose regression for each point, we select a single 6D pose for each object among the candidates, taking the semantic labels into account.  It is not desirable to take a mean of candidates naively, which can harm the orthonormality of estimated 3D rotation matrix. For each instance, we select the final pose by converting rotation matrices to rotation vectors, computing mean of the vectors, and choosing the candidate with the smallest Euclidean distance to the mean. The same process can be applied to the translation vectors as well.

\subsubsection{Disentangled loss functions for 6D pose}
\par We disentangled 6D pose loss functions for 3D rotation and 3D translation respectively . While 3D translation vectors are supervised with a loss function based on L1 norm, the supervision on the parameters for 3D rotation matrix should be treated carefully. There are multiple ground truths for symmetric objects, so we separated the case for symmetric and asymmetric objects following \cite{xiang2017posecnn} utilizing the rotation matrix we selected.
\begin{align}
    &\mathcal{L}_t = \sum_c\sum_{n=1}^N ||t_n - t_n^*||_1 {I}(\mathcal{C}_n=c)  \\
    &\mathcal{L}_R = 
    \begin{cases}
        {L}_{R, asym} &\textit{if asymmetric} \\ 
        {L}_{R, sym} &\textit{if symmetric}
    \end{cases} \\
    & \textit{where} \ 
    \begin{cases}
   R_c = \underset{R_i\in \mathcal{R}_c}{argmin}||R_i - \bar{\mathcal{R}}_c||_2\\ \nonumber
       \mathcal{L}_{R, asym}= \underset{c}{\sum} \underset{\mathrm{x}\in \mathcal{M}_c}{avg}||R_c\mathrm{x} - R_c^*\mathrm{x}||_2 \\ \nonumber
    \mathcal{L}_{R, sym}= \underset{c}{\sum} \underset{\mathrm{x_1}\in \mathcal{M}_c}{avg}\left[\underset{\mathrm{x_2}\in \mathcal{M}_c}{min}||R_c\mathrm{x_1} - R_c^*\mathrm{x_2}||_2\right]
    \end{cases} \\ \nonumber
\end{align}
where $t_n$ is candidate 3D translation and ${I}(\mathcal{C}_n=c)$ denotes indicator function that class of point n equals to class c. $R_c$ is the selected rotation matrix for each class among the candidates $\mathcal{R}_c $ where $\bar{\mathcal{R}}_c$ is the mean of them, and $\mathcal{M}_c$ is a set of vertices of mesh model of class $c$.

\par The overall loss for KGpose to supervise the end-to-end 6D pose estimation process through multi-task learning is then:
\begin{align}
\mathcal{L}_{total}
=\alpha\mathcal{L}_{seg}
+\beta\mathcal{L}_{kp}+\gamma\mathcal{L}_{geom}
+\delta(\mathcal{L}_{R}+\mu\mathcal{L}_{t}).
\end{align}

\par Most end-to-end pose estimation frameworks utilize object detection algorithm ahead of pose regression\cite{wang2021gdr, di2021so, lin2022e2ek, cao2022dgecn} to localize single object. Meanwhile, our approach doesn't require localization but leverages the candidates about 6D pose and learn to aggregate them to select 6D pose for each instance, which enables pose estimation for multiple instances at once.

\begin{table*}[t]
    \centering
    \begin{tabular}{l|c|c|c|c|c|c|c|c|c|c} 
        \hline
        \multicolumn{1}{c|}{\multirow{2}{*}{Object}} & \multicolumn{2}{c|}{PoseCNN\cite{xiang2017posecnn}} & \multicolumn{2}{c|}{DenseFusion\cite{wang2019densefusion}} & \multicolumn{2}{c|}{PVN3D\cite{he2020pvn3d}} & \multicolumn{2}{c|}{FFB6D\cite{he2021ffb6d}} & \multicolumn{2}{c}{KGpose (Ours)} \\\cline{2-11}
        & ADDS& ADD(S) & ADDS& ADD(S) & ADDS& ADD(S) & ADDS& ADD(S) & ADDS& ADD(S) \\
         \hline
        002 master chef can & 83.9& 50.2& 95.3& 70.7& 96.0& 80.5& 96.3& 80.6 & 95.3 & 80.8 \\
        003 cracker box & 76.9& 53.1& 92.5& 86.9& 96.1& 94.8& 96.3& 94.6 & 95.3 & 92.2 \\
        004 sugar box & 84.2& 68.4& 95.1& 90.8& 97.4& 96.3& 97.6& 96.6 & 96.6 & 93.9 \\
        005 tomato soup can & 81.0& 66.2& 93.8& 84.7& 96.2& 88.5& 95.6& 89.6 & 95.7 & 84.4 \\
        006 mustard bottle & 90.4& 81.0& 95.8& 90.9& 97.5& 96.2& 97.8& 97.0  & 98.2 & 95.8 \\
        007 tuna fish can & 88.0& 70.7& 95.7& 79.6& 96.0& 89.3& 96.8& 88.9  & 98.1 & 93.6 \\
        008 pudding box & 79.1& 62.7& 94.3& 89.3& 97.1& 95.7& 97.1& 94.6  & 98.1 & 95.5\\
        009 gelatin box & 87.2& 75.2& 97.2& 95.8& 97.7& 96.1& 98.1& 96.9  & 97.0 & 94.9\\
        010 potted meat can & 78.5& 59.5& 89.3& 79.6& 93.3& 88.6& 94.7& 88.1  & 94.5 & 87.5\\
        011 banana & 86.0& 72.3& 90.0& 76.7& 96.6& 93.7& 97.2& 94.9 & 97.4 & 94.0\\
        019 pitcher base & 77.0& 53.3& 93.6& 87.1& 97.4& 96.5& 97.6& 96.9 & 96.8 & 94.4\\
        021 bleach cleanser & 71.6& 50.3& 94.4& 87.5& 96.0& 93.2& 96.8& 94.8 & 96.7 & 94.0\\
        \textbf{024 bowl} & 69.6& 69.6& 86.0& 86.0& 90.2& 90.2& 96.3& 96.3 & 92.7 & 92.7\\
        025 mug & 78.2& 58.5& 95.3& 83.8& 97.6& 95.4& 97.3& 94.2 & 96.2 & 91.4\\
        035 power drill & 72.7& 55.3& 92.1& 83.7& 96.7& 95.1& 97.2& 95.9 & 97.4 & 95.0\\
        \textbf{036 wood block} & 64.3& 64.3& 89.5& 89.5& 90.4& 90.4& 92.6& 92.6 & 91.0 & 91.0\\
        037 scissors & 56.9& 35.8& 90.1& 77.4& 96.7& 92.7& 97.7& 95.7 & 95.2 & 90.2\\
        040 large marker & 71.7& 58.3& 95.1& 89.1& 96.7& 91.8& 96.6& 89.1 & 92.1 & 84.6\\
        \textbf{051 large clamp} & 50.2& 50.2& 71.5& 71.5& 93.6& 93.6& 96.8& 96.8 & 95.6 & 95.6\\
        \textbf{052 extra large clamp} & 44.1& 44.1& 70.2& 70.2& 88.4& 88.4& 96.0& 96.0 & 94.5 & 94.5\\
        \textbf{061 foam brick} & 88.0& 88.0& 92.2& 92.2& 96.8& 96.8& 97.3& 97.3 & 97.2 & 97.2\\\hline
        \multicolumn{1}{c|}{MEAN} & 75.8& 59.9& 91.2& 82.9& 95.5& 91.8& 96.6& 92.7 & 95.7 & 92.1\\\hline
    \end{tabular}
    \caption{Quantitative results on YCB-Video Dataset according to ADD-S AUC and ADD(S) AUC metrics. Names of symmetrical objects are in bold }
\end{table*}

\begin{figure*}[t!]
    \centering
    {\includegraphics[width=0.24\textwidth]{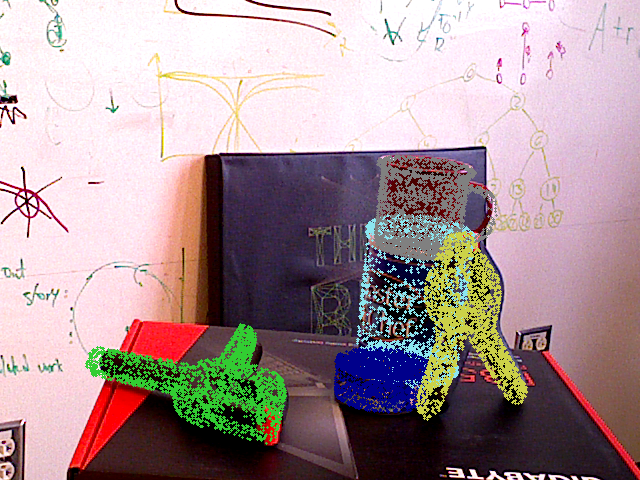}}
    {\includegraphics[width=0.24\textwidth]{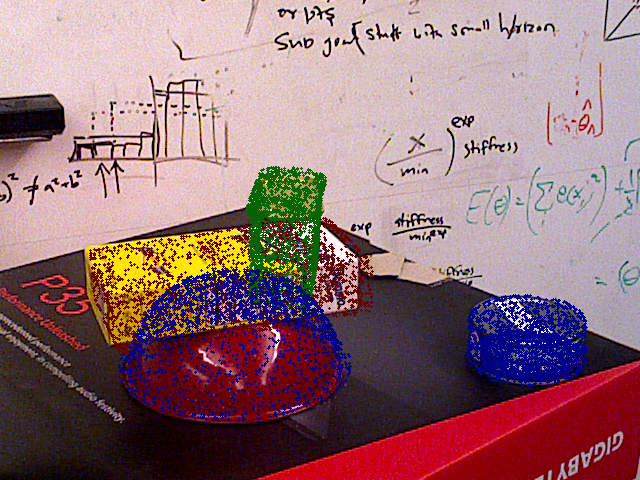}}
    {\includegraphics[width=0.24\textwidth]{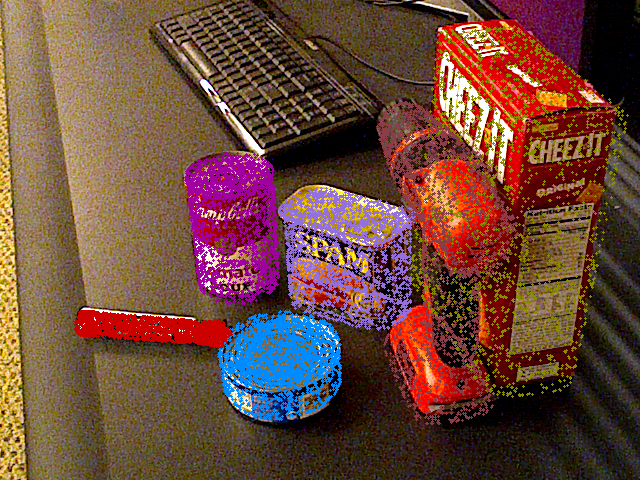}}
    {\includegraphics[width=0.24\textwidth]{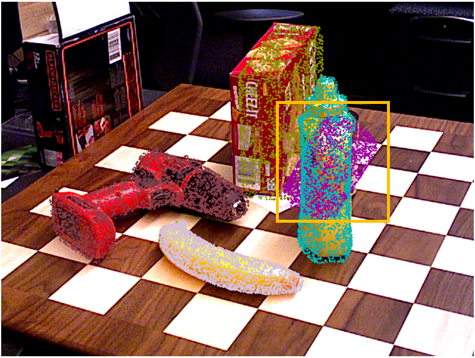}}
    \caption{Qualitative results on YCB-Video Dataset. Vertices of object model are transformed by estimated pose and projected to the 2D image to visualize the performance on 6D pose estimation. Our model works well but still has difficulty on heavily occcluded scenes (see the right figure).}
    \label{fig:3.2}
\end{figure*}


\section{Experiment}

\subsection{Training and Implementation}
\par All experiments are implemented using PyTorch. We trained our network for 40 epochs  with a batch size of 4 using four NVIDIA RTX 3090 GPUs. We adopted AdamW optimizer\cite{loshchilov2017decoupled} and one-cycle learning rate scheduler\cite{smith2019super}. We set $\alpha=2.0, \beta=1.0, \gamma=1.0, \mu=2.0$ and $\delta=0.01$ for 20 epochs and $\delta=1.0$ for the rest of the epochs.

\par To extract features from point cloud efficiently, we randomly sampled 12800 points from depth images of size 480 $\times$ 640 after filling the holes\cite{ku2018defense} and utilized it as input for RandLA-Net\cite{hu2020randla}.

\subsection{Datasets}
\par We evaluated our network on YCB-Video dataset, which contains 92 RGB-D image sequences about the subset of 21 YCB objects providing semantic masks as well as 6D pose of objects in each scene. This dataset is challenging due to image noise, low-texture, high occlusion and various shapes including symmetric objects. Following \cite{xiang2017posecnn},  the dataset is split into 80 videos for training and remaining 12 videos for testing. Synthetic images of the rendered YCB objects are provided and also used for training.

\subsection{Metrics for Evaluation}
\par We utilized two common metrics ADD and ADD-S to evaluate 6D pose estimation. For asymmetric objects, the ADD metric measures whether the average deviation of transformed model vertices is within 10\% of the object's diameter. For symmetric objects, the ADD-S metric is used, which calculates the average distance to the closest point, considering the minimum of all point-to-point distances between the transformed model and the ground truth. .
\begin{align}
    &\text{ADD} = \frac{1}{m} \sum_{v \in M} \|(Rv + T) - (R^*v + T^*)\|  \\ 
    &\text{ADD-S} = \frac{1}{m} \sum_{v_1 \in M} \underset{v_2 \in M}{min}\|(Rv_1 + T) - (R^*v_2 + T^*)\|
\end{align}
where $R, T$ are predicted rotation and translation and $R^*, T^*$ are ground-truths of them.

\par During evaluation, we computed the ADD-S AUC (Area Under Curve) and ADD(S) AUC by varying the distance threshold following \cite{xiang2017posecnn, wang2019densefusion}. Especially, ADD(S) AUC adopts ADD distance for asymmetric objects and ADD-S distance for symmetric ones.

\subsection{Evaluation Results}
\par Our KGpose achieved mean scores of 95.7\% for ADD(S) and 92.1\% for ADD, demonstrating competitive performance compared to state-of-the-art methods like PVN3D\cite{he2020pvn3d} and FFB6D\cite{he2021ffb6d}. KGpose performed well on both symmetric and asymmetric objects, showing particular strength on challenging items such as the mustard bottle and tuna fish can. Qualitative results indicated precise performance in various scenarios, though some difficulties persisted in heavily occluded scenes.

\section{Conclusion} \label{sec:Conclusion}
\par This paper introduced KGpose, an end-to-end framework for object 6D pose estimation. KGpose combines keypoint-based methods with learnable direct pose regression using a `keypoint-graph' representation. Our approach demonstrates competitive performance on the YCB-Video dataset, suggesting point-wise pose voting scheme to enable multi-object pose estimation without extra object localization algorithm.

\par Directions for future researches would include extending KGpose to handle a wider variety of objects and challenging scenarios like outdoor environment, or integrating with downstream robotic tasks. Exploring self-supervised learning with less supervision or addressing novel objects that have never been seen would also be desirable. These improvements will enhance the applicability of 6D pose estimation in real-world robotic scenarios.

\ifCLASSOPTIONcaptionsoff
  \newpage
\fi

\bibliographystyle{IEEEtran}
\bibliography{references}

\vspace{-1.5cm}

\end{document}